\documentclass[runningheads]{llncs}
\usepackage[T1]{fontenc}
\usepackage{graphicx}
\usepackage{adjustbox}
\usepackage{makecell}
\usepackage{subcaption}
\usepackage{svg}  
\usepackage[acronym]{glossaries}
\usepackage{booktabs}
\usepackage{tcolorbox}
\usepackage{xcolor}
\usepackage{tabularx}

\newacronym{AI}{AI}{Artificial Intelligence}
\newacronym{GTD}{GTD}{Generated Text Detection}
\newacronym{IPAD}{IPAD}{Inverse Prompt for AI Detection}
\newacronym{LLM}{LLM}{Large Language Model}
\newacronym{roberta}{RoBERTa}{Robustly optimized BERT approach}
\begin{document}

\title{Sure! Here's a short and concise title for your paper: ``Contamination in Generated Text Detection Benchmarks''}
\titlerunning{Contamination in Generated Text Detection Benchmarks}
%
\author{Philipp Dingfelder\orcidID{0009-0000-2038-2489} \and
Christian Riess\orcidID{0000-0002-5556-5338}} 
\authorrunning{P. Dingfelder, C. Riess}
\institute{IT Security Infrastructures Lab, FAU Erlangen-N{\"u}rnberg, Martensstr. 3, 91058 Erlangen, Germany \\
\email{\{philipp.dingfelder,christian.riess\}@fau.de}}
\maketitle              
\begin{abstract}
Large language models are increasingly used for many applications. To prevent illicit use, it is desirable to be able to detect AI-generated text. Training and evaluation of
such detectors critically depend on suitable benchmark datasets. Several groups took on the tedious work of collecting, curating, and publishing large and diverse datasets for this task.
However, it remains an open challenge to ensure high quality in all
relevant aspects of such a dataset. For example, the DetectRL benchmark exhibits
relatively simple patterns of AI-generation in 98.5\% of
the Claude-LLM data. These patterns may include introductory words such as
``Sure! Here is the academic article abstract:'', or instances where the LLM
rejects the prompted task.

In this work, we demonstrate that detectors trained on such data use such patterns as shortcuts, which facilitates spoofing attacks on the trained detectors. We consequently reprocessed the DetectRL dataset with several cleansing operations. Experiments show that such data cleansing makes direct attacks more difficult.
The reprocessed dataset is publicly available.

\keywords{Generated Text Detection \and Data Quality \and Benchmarking \and DetectRL}\newline

\end{abstract}

\section{Introduction}
Generative \gls{AI} can assist with various tasks, including programming, creating media and content, and utilising chatbots as an alternative to traditional internet searches.
However, alongside these positive applications, the potential misuse of generative \gls{AI} is also well known. These include plagiarism in educational settings~\cite{Pudasaini.2024} and the creation of deepfakes~\cite{Mubarak.2023}. In particular, the text domain attracts a high level of attention due to the increasing capabilities of \glspl{LLM}. For instance, recent reasoning models ~\cite{anthropic.2025,opnenai.2025} have become more practical in scientific research~\cite{QileJiang.2025}. 
This, combined with the unreliability of humans as detectors~\cite{Gao.2023,Gehrmann.2019}, has resulted in a growing number of automatic detectors and \gls{GTD}-Benchmarks, making it an extremely active area of research.

One family of detectors works with an actively embedded signal, like a watermark. They usually require control of the text generation process for embedding the watermark,
which is only given for providers of \gls{LLM} systems~\cite{SoumyaSuvraGhosal.2023}.
Another family of detectors work passively. They forensically examine a sequence of text for traces of a generator. Passive detectors are by tendency less reliable, but they have the advantage that they can be flexibly applied post-hoc and without control of the generation process.
Within this family, even more flexibility can be achieved if detectors have the ability to generalise across different types of generators. This brings the benefit that the detector is not limited to the exact models in the training set when testing it on text in the field~\cite{HansSCKSGGG24}.

Even if a generator generalizes well, it is still evident that its overall quality critically depends on the quality of its training and calibration data.
This particularly affects the dectector's ability to generalise to various real-world scenarios~\cite{Pudasaini.2024}.
DetectRL and related benchmarks aim to provide such data~\cite{DBLP:conf/acl/WangMISSTAM0AAH24,wang-etal-2024-semeval-2024,DBLP:conf/nips/WuZWY0YC24}.
They provide samples of various attacks from multiple LLMs with the goal of achieving a high degree of realism~\cite{DBLP:conf/nips/WuZWY0YC24}.
However, our investigation on DetectRL reveals that data quality may not always be adequately addressed.
The benchmark data contains several artifacts of LLM-generated content, such as \gls{LLM}-typical starting phrases like 
\textit{``Sure! Here is the abstract for the given title: ...''} or rejection phrases such as 
\textit{``I apologize, upon further reflection I do not feel comfortable generating fictional negative reviews.''}
While the first type of artifacts may cause trained detectors to learn a shortcut, multiple rejections may lead to repeating rejection phrases that are easy to detect. If not excluded in the evaluation, these phrases may therefore lead to overly optimistic performance metrics.

The DetectRL dataset is gaining popularity within the research community. However, to the best of our knowledge,  neither the original work nor follow-up works report or discuss these issues. 
With this work, we hope to close this gap. We analyse the extent and potential impact of these issues. 
In order to improve the usefulness of the DetectRL dataset, we also provide a reprocessed
version of the DetectRL benchmark dataset for download\footnote{\url{
https://www.cs1.tf.fau.de/research/multimedia-security/code/reprocessed-detectrl-dataset/}}.

The paper is organized as follows. Section~2 reviews the state of the art in \gls{GTD} methods.
Section~3 investigates the contamination in the DetectRL benchmark. Section~4 describes 
methods to estimate the impact of the contamination and the reprocessing of the dataset. Section~5 reports experimental results on the contamination and the reprocessed dataset. Section~6 concludes this work.

\section{Related Work}
The field of generated text detection is a fast growing branch of research. The following literature review can hence only cover a small portion of related works. We refer to Ghosal~\textit{et al.}~\cite{SoumyaSuvraGhosal.2023} and Wu~\textit{et al.}~\cite{DBLP:journals/coling/WuYZYCW25} for more elaborate surveys on the topic. 

One family of \gls{GTD} methods involves active detection schemes with an embedded signal. 
Arguably the most relevant subcategory makes use of watermarks that either distort the output probabilities of the tokens \cite{Kirchenbauer.2023}, or use a pseudo-random generator~\cite{ScottAaronson.February222023}.
Another group of active detection schemes are retrieval-based methods~\cite{krishna2023paraphrasing}. However, active detection schemes all share the limitation that they require access to the text generation process, which is in many application scenarios unavailable.

Such access to the generation process is not necessary for passive detectors. These form a second family of methods that can be applied post-hoc to probe any text for its origin.
According to Ghosal~\textit{et al.}~\cite{SoumyaSuvraGhosal.2023}, passive detectors can be coarsely categorised into trained detectors and zero-shot detectors.
Trained detectors like \gls{roberta} are based on pre-trained transformers usually with an additional classification head~\cite{DBLP:journals/corr/abs-1907-11692}. 
In contrast, zero-shot detectors do not require any generator-specific training. Therefore, they are considered to be more model-agnostic. They probe specific properties of the text to distinguish model-generated text from human-written text.
Initial attempts such as GLTR are based on the observation, that language models only use a limited subset of the distribution of natural language with a high likelihood~ \cite{Gehrmann.2019}.
A subsequent work uses the perplexity as a measure of naturalness of the text~\cite{DBLP:journals/corr/abs-2305-18226,DBLP:conf/iclr/Yang0WPWC24}.
Binoculars extends this work by using cross-perplexity from a reference
model~\cite{HansSCKSGGG24}.
This approach ensures that traces of generated text can also be found in text stemming from particular prompts that make the output appear less likely to be \gls{LLM}-generated in the first place. 
\gls{IPAD} is a recent zero-shot detector that uses a prompt inverter to regenerate text similar to the input text~\cite{DBLP:journals/corr/abs-2502-15902}. This is followed by an distinguisher module that is used to classify the text based on the alignment of the input and the regenerated text. 

Besides these recent advances in generated text detection, there have also various attacks been published that challenge the practical effectiveness of \gls{GTD} methods. 
Paraphrasing is arguably the most prominent attack to evade detection. It can be implemented in different ways. For example, DIPPER is a language model that has been specifically trained for paraphrasing~\cite{krishna2023paraphrasing}. Other possibilities are to use a general-purpose \gls{LLM}, or to perform paraphrasing via back-translation~\cite{DBLP:conf/nips/WuZWY0YC24}.
Additionally, humans themselves may also paraphrase the generated text. 

Beyond basic paraphrasing attacks, it has been demonstrated that a sequence of multiple paraphrasing steps makes the detection of LLM-generated text notably harder~\cite{HanlinZhang.2024}.
Further scenarios include polishing with another LLM, mixing of human-written and \gls{LLM}-generated content, employing various prompting strategies, and launching adversarial attacks such as incorporating spelling errors \cite{DBLP:conf/nips/WuZWY0YC24}.

There are several datasets to benchmark the detection of generated text. Widely used datasets besides DetectRL are for example
M4GT-Bench~\cite{DBLP:conf/acl/WangMISSTAM0AAH24}  or SemEval-2024 Task 8~\cite{wang-etal-2024-semeval-2024}. All of these datasets cover multiple application domains like reviews, abstracts, creative writing, and news articles. M4GT-Bench and SemEval-2024 offer data in multiple languages.
DetectRL only contains English-language text. However, it stands out in the number of provided attacks, including different variants of paraphrasing attacks, adversarial attacks, and polishing attacks. 

\section{Analysing the Level of Contamination}

Trained detectors may use simple syntactic features as shortcuts for detection. For example, 
Doughman~\textit{et al.} shows that detectors can pick up 
patterns in punctuation or whitespaces if the dataset exhibits some form of bias there~\cite{doughman-etal-2025-exploring}.
When visually sifting through the dataset, we noticed some hints of contamination, like redundancies and typical phrases of generative chat models that are not directly relevant for the task at hand. We subsequently started to search for these patterns more thoroughly and analyse the overall volume of the contamination in the benchmark. To this end, we 
visually inspected a representative subset of the data in order to gain a qualitative understanding of the types of the contamination.
In a second step, we crafted regular expressions that were derived
from that inspection to also obtain a quantitative number of contamination cases.

The design of the regular expressions followed observations from the dataset. The observations are described in the next paragraphs and illustrated in Table~\ref{tab:reg_exp_cleaning}.

\noindent \textbf{Pattern Rejection.} As the \glspl{LLM} used are instruction-tuned to align with ethical guidelines and prevent misuse, rejections of the \glspl{LLM} are possible. These typically result in \gls{LLM}-dependent standard phrases, which were incorporated into the first group of regular expressions, which we call Pattern Rejection (or ``rejection'' for short). 

\noindent \textbf{Prompt-specific Pattern.} For two types of tasks (polishing human- or \gls{LLM}-generated content and SICO prompting), some \gls{LLM} responses repeat keywords from the prompt or provide a summary of their actions. These task-specific responses represent the second category, which we call Prompt-specific Pattern (or ``prompt'' for short). 

\noindent \textbf{Pattern Beginning.} The third type of pattern is arguably the one that is visually most easily detected. It captures instances where the response begins with ``Here is...'' or a similar phrase. This pattern is especially prevalent in the Claude version used to generate the benchmark, occurring in 94.7\% of cases of Claude-generated text. 
We call this type Pattern Beginning (or ``beginning'' for short).

\noindent \textbf{Domain-specific Pattern.} Domain-specific patterns are closely related to the task at hand such as keywords like ``abstract'' for tasks that generate arXiv abstracts or ``reviews'' for tasks that generate Yelp reviews. However, these criteria are somewhat unsharp. When used as a selection criterion in regular expressions, then it may be the case that phrases are wrongly selected as false positives. Further below in the counting of pattern occurrences in Tab.~\ref{tab:contamination}, we did not normalize for these false positives, but in the reprocessed dataset, we explicitly double-checked these cases to remove false positives. In any case, we refer to this type as Domain-specific Pattern (or ``domain'' for short).

\noindent \textbf{Assistant Pattern.} The fifth type of pattern is a specific feature that depends on the combination of task and LLM. We particularly observed this for PaLM.
For example, PaLM rejected 396 out of 700 creative writing examples when it was using few-shot prompting for generation.
One reason might have been that human-written examples that were used for few-shot prompting might have contained (allegedly) harmful content.
However, such strict rejections do not occur with other LLMs. In addition, either system-related messages that are not intended for the user or the phrase '[assistant]:' are provided before the answer. We refer to this pattern as Assistant Pattern (or ``assistant'' for short).

\begin{table}[tb]
    \centering
    \caption{Regular expressions for data analysis and cleaning. Some long patterns have been omitted for reasons of space and clarity. }
    \label{tab:reg_exp_cleaning}
    \begin{tabular}{lp{0.7\textwidth}}
        \toprule
        \textbf{Pattern Variable} & \textbf{Regular Expression} \\
        \midrule
        \texttt{1. Pattern Rejection} & \texttt{\detokenize{(.*I apologize, upon further reflection.*)|(.*((only)|(just)) a language model.*)|...}} \\
        \midrule
        \multicolumn{2}{l}{2. Prompt-specific Pattern:} \\
        \texttt{SICO-Prompting} & \texttt{\detokenize{(in a human\s?\w{0,20}\s?style)}} \\
        \texttt{Polishing} & \texttt{\detokenize{(grammar[\w\s,]{1,40}spelling)|...}} \\
        \midrule
        \texttt{3. Pattern Beginning} & \texttt{\detokenize{(Voici un|Here is|Here are|Here's|Sure[,!]?\s?here)}} \\
        \midrule
        \multicolumn{2}{l}{4. Domain-specific Pattern} \\
        \texttt{Pattern Article} & \texttt{\detokenize{(given article title|provided article title)}} \\
        \texttt{Pattern Yelp Review} & \texttt{\detokenize{(review's first sentence|review)}} \\
        \texttt{Pattern Arxiv Abstract} & \texttt{\detokenize{(abstract|academic article)}} \\
        \texttt{Pattern XSUM} & \texttt{\detokenize{(article)}} \\
        \midrule
        \texttt{5. Assistant Pattern} & \texttt{\detokenize{(?:.*)(((\[system\])|(\[user\])|(\[assistant\]))\s*\w{0,20}|(\*\*assistant))([:]?[\*]{2}|[:])}} \\
        \bottomrule
    \end{tabular}
\end{table}

A quantitative overview of the number of contaminations across the four LLMs in DetectRL is shown in Table~\ref{tab:contamination}. 
Selection of entries is applied without replacement, i.e., if an entry matches a regular expression, then it is not further analysed with other regular expressions. 
Here, ChatGPT shows overall very few contamination, most notably are a few cases
of domain-specific patterns.
Claude-instant exhibits most cases of contamination across all LLMs, and the big majority of contaminations occurs in the Pattern Beginning with $13,261$ cases. The other notable pattern in Claude-instant are $448$ cases of Pattern Rejection. Google-PaLM and Llama-2-70b are also affected in a non-negligible number of cases, but nevertheless to a much smaller extend than Claude-instant. Both exhibit around $1300$ to $1500$ cases of Pattern Beginning. Google-Palm accumulates an additional total of around $2000$ cases of Pattern Rejection, Domain-specific Patterns, and Assistant Pattern, whereas Llama-2-70b exhibits additional $520$ cases of Prompt-specific Pattern.

\begin{table}[tb]
\setlength{\tabcolsep}{4pt} 
\centering
\caption{Distribution of the identified types of contaminations that were found with regular expressions in DetectRL.
 The total number of potentially contaminated text found with regular expressions is 20,325 (total entries: 56,000).}
\label{tab:contamination}
\begin{tabular}{rlll}
\toprule
\textbf{} & \textbf{llm\_type} & \textbf{category} & \textbf{count} \\
\midrule
\midrule
0  & ChatGPT        & rejection & 6   \\
1  & ChatGPT        & prompt    & 9   \\
2  & ChatGPT        & beginning & 8   \\
3  & ChatGPT        & domain    & 264 \\
\midrule
4  & Claude-instant & rejection & 448   \\
5  & Claude-instant & prompt    & 79    \\
6  & Claude-instant & beginning & 13,261 \\
7  & Claude-instant & domain    & 8     \\
\midrule
8  & Google-PaLM    & rejection & 1,703  \\
9  & Google-PaLM    & prompt    & 38    \\
10 & Google-PaLM    & beginning & 1,585  \\
11 & Google-PaLM    & domain    & 457   \\
12 & Google-PaLM    & assistant & 469   \\
\midrule
13 & Llama-2-70b    & rejection & 27    \\
14 & Llama-2-70b    & prompt    & 520   \\
15 & Llama-2-70b    & beginning & 1,352  \\
16 & Llama-2-70b    & domain    & 91    \\
\bottomrule
\end{tabular}
\end{table}

\section{Methodology and Data Cleansing}
The quantitative examination of contamination showed that 
the dataset contains a large number of generator-typical phrases. This raises
the question whether this contamination influences trained detectors.
We investigate this question with the \gls{roberta}-Base detector, which is the best performing detector in the DetectRL paper. In a second step, we reprocess the dataset to remove the identified contaminations, and re-evaluate the \gls{roberta} detector.

\subsection{Impact of the Contamination}
The contaminations occur similarly in the training and in the test data of the DetectRL benchmark.
Hence, the guiding hypothesis in this experiment is that \gls{roberta} learns these contaminations during training, and consequently predicts a higher likelihood of \gls{AI}-generation on test data that contains the same contaminations. 
%
To investigate this hypothesis, we use the post-hoc explainability method
SHAP~\cite{lundberg2017unified}, similar to Doughman~\textit{et
al.}~\cite{doughman-etal-2025-exploring}. The idea is to study the impact of
specific tokens on the outcome of a prediction. If the model learns a shortcut
from a contamination, then one can expect to identify the contaminated tokens as
particularly important components in the prediction.

The \gls{roberta} model is trained on a single \gls{LLM} multi-domain dataset, as most types of contamination seem to exhibit LLM-specific traces.
SHAP is applied to the outputs of this \gls{roberta} model. 
Then, we also perform adversarial attacks analogous to \textit{spoofing} in watermarks \cite{sadasivan2025can}. Here,  we try to lure the detector into misclassifying a human-written text as \gls{AI}-generated
by appending the phrase \textit{'Here is a 7 sentence abstract for the provided article title: '}.
The impact of this attack is analyzed by studying the changes in classification probabilities.
Finally, we analyse the generalisation capability of the uncleaned dataset compared to the cleaned dataset as a first step to determine whether contamination would impact the results reported by the DetectRL Benchmark.

\subsection{Removal of the Identified Contaminations}
For the removal of the patterns, we use the same regular expressions that were also used to detect and analyse the contamination. 
Rejections by the LLM typically involve predefined standard clauses and are easy to identify. To create a meaningful and challenging benchmark, we completely remove them from the dataset and replace them with null values.
The other contaminants are typically introductory statements or, less frequently, explanations provided by the LLM at the end of the task. We therefore remove the entire sentence, starting with the contaminant and ending with the next sentence-ending punctuation mark (‘?’, ‘!’ or ‘.’) or the next colon.
There is a uniform prompt for each domain, with only the context variable (e.g. the title for which an abstract is to be generated, and the required number of sentences). The task and prompt itself (e.g. generating an abstract) remains the same for an individual domain.
Consequently, the individual LLMs tend to have a repetitive structure. This can easily be corrected using the pattern-matching procedure described above. Nevertheless, two types of error can occur that need to be corrected. Either too little is removed, leaving the contamination partially present, or too much is removed, rendering the text unusable.

In the former case, a \gls{roberta} model is trained on contaminated text. This model is then used to classify the processed samples. Samples that are most likely to be generated by \gls{AI}, even after cleaning, are likely to still exhibit contamination and are marked for a further processing step.

To address cases where the removal was too rigorous, we first collect samples with a low absolute number of tokens.
Since \glspl{LLM} are prompted to generate text with the same number of sentences as the analogous human-written example, they should produce a similar number of sentences and tokens.
Therefore, we also re-clean samples where there are large relative differences in the number of sentences and tokens between the processed \gls{LLM} entry and the corresponding human-written text.

The re-cleaning of the selected samples is performed using \textit{gpt-4.1-mini} with few-shot prompting. The model is instructed to remove only the contamination and return the uncontaminated text. The answer to the original task, and therefore the traces of the original LLM, should remain unchanged. To ensure the intended behaviour, we follow OpenAI's associated prompting guide \cite{maccallum2025gpt41}.

\begin{table}[tb]
\caption{Example of a model-generated abstract with a \gls{LLM} typical beginning. Used in Figure \ref{fig:shap_normal} to determine the SHAP values of contaminated text. }
\label{A1:contaminated_text}
\begin{tabular}{p{\textwidth}}
\toprule
\textit{``Here is a 10 sentence abstract for the article title "Fundamental Limits to Position Determination by Concentration Gradients":Organisms across nature have evolved to determine their position using concentration gradients of signaling molecules. However, gradient sensing poses fundamental physical limitations in accuracy and precision. This study explores the biophysical limits to localization from concentration gradients. A computational model is developed to describe gradient formation and interpret signal transduction by cell surface receptors in response to different gradient profiles. The model accounts for stochastic variability in ligand-receptor binding and finite numbers of receptors. Spatial resolution is shown to degrade significantly for shallow gradients and small cell sizes due to stochastic noise. Optimal gradient shapes are identified that allow sub-cellular precision even in noisy conditions. Experimental measurements of gradient sensing match predictions from the biophysical model. The results establish baselines for position determination based solely on gradient interpretation. Fundamental tradeoffs between accuracy, precision and measurement time are characterized. This provides insights into the physical design constraints shaped by evolution in gradient-based navigation across scales from microbes to multicellular organisms.''}
- LLM: Claude-instant, domain: Arxiv abstracts \\
\bottomrule
\end{tabular}
\end{table}

\section{Results}

First, the prediction of a model trained on contaminated data is explained, followed by one trained on data cleaned using only the regular expressions.
All experiments are performed on a model trained on the Claude multi-domain dataset, as this dataset has been shown to be the most contaminated (cf. Tab. \ref{tab:contamination}).

\begin{figure}[tb]
    \centering
    \begin{subfigure}[t]{0.48\textwidth}
        \centering
        \includegraphics[width=\linewidth]{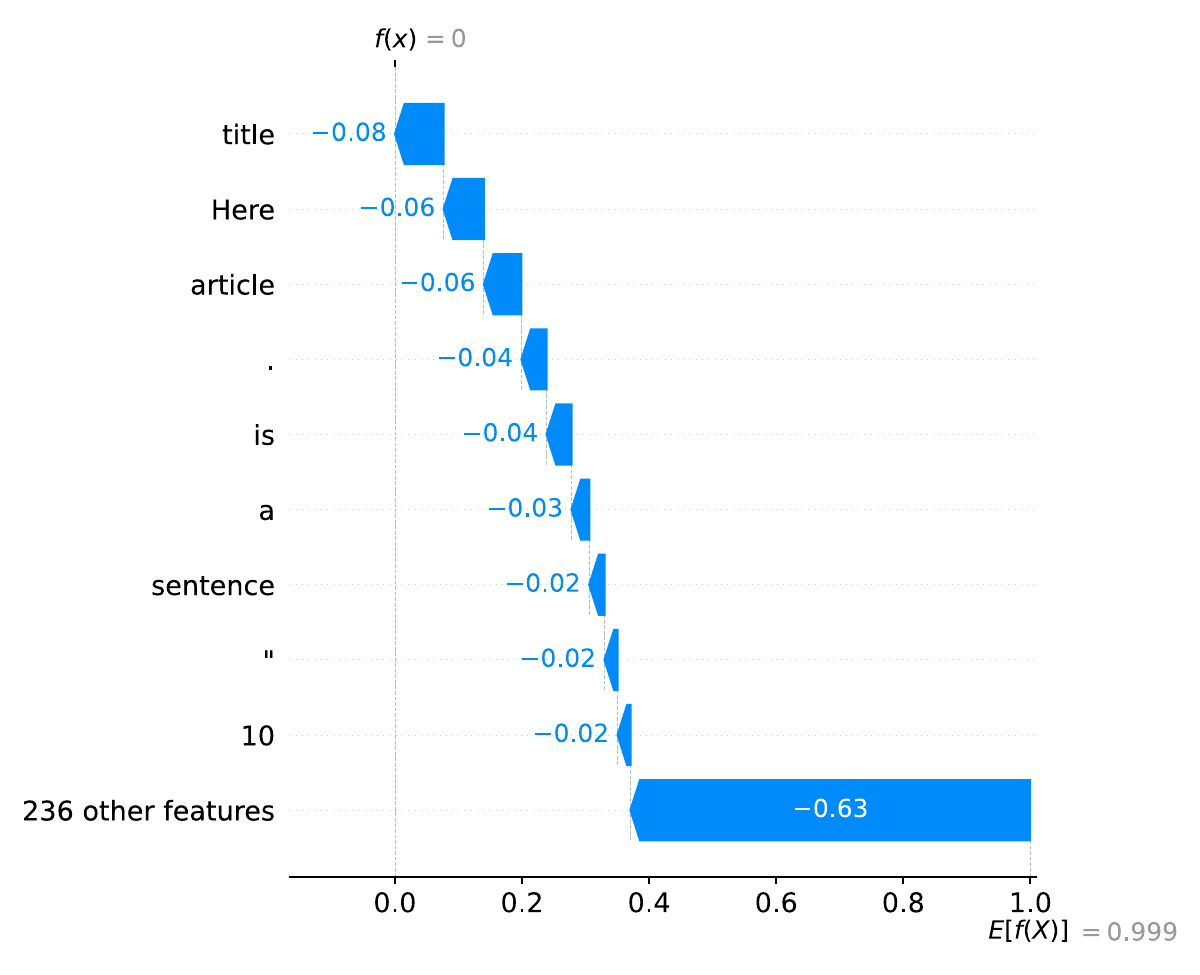}
        \caption{Using the original DetectRL Benchmark data.}         \label{fig:shap_normal_a}
    \end{subfigure}
    \hfill
    \begin{subfigure}[t]{0.48\textwidth}
        \centering
        \includegraphics[width=\linewidth]{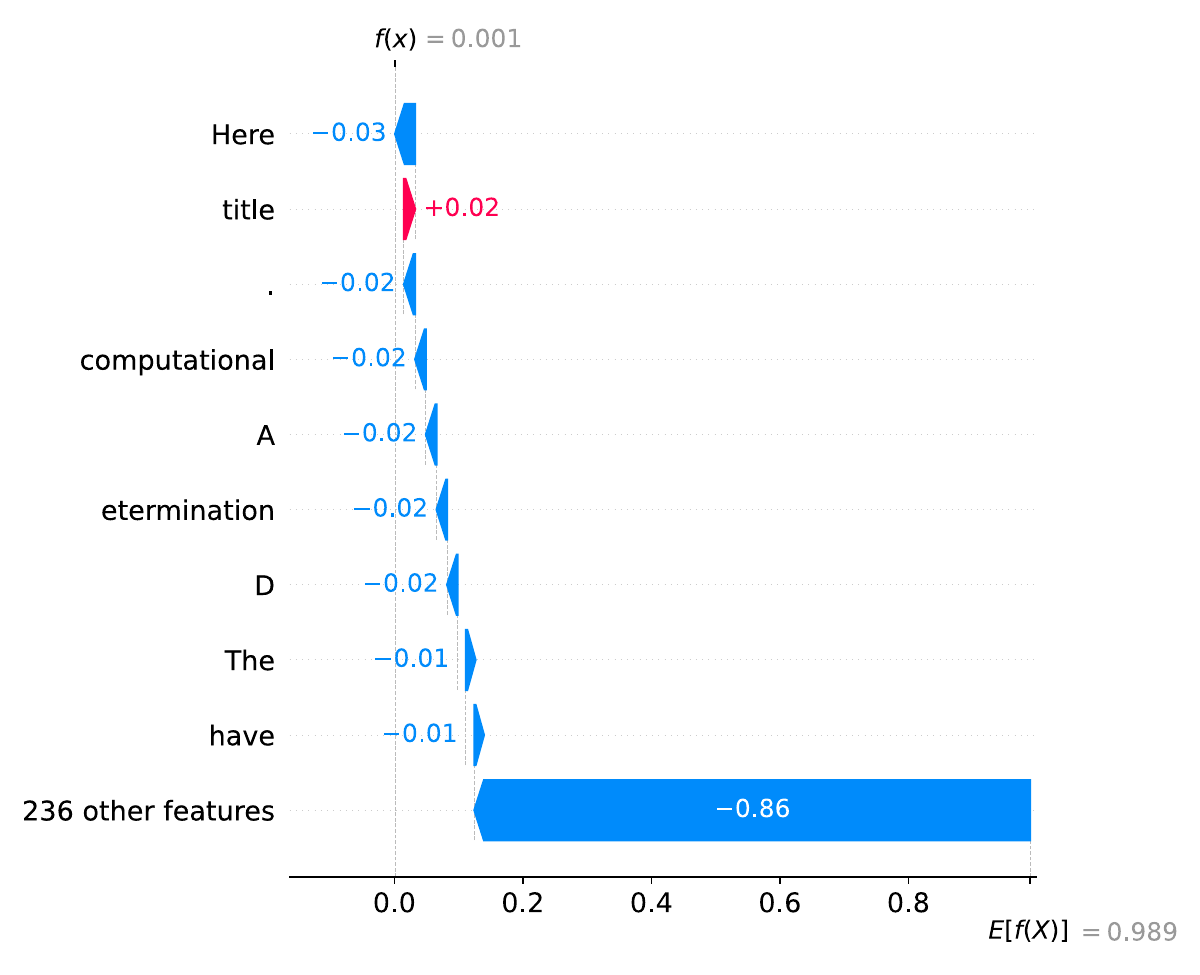}
        \caption{Using data re-cleaned with regular expressions.}
        \label{fig:shap_normal_b}
    \end{subfigure}
    \caption{Comparison of SHAP token importance of a text for a model trained on Claude-generated, multi-domain data. The sample is labeled as LLM-generated. The classifier is trained to predict whether text is human-generated or AI-generated. Values close to 1 indicate human-generated text, while values below 0.5 indicate AI-generated text.}\label{fig:shap_normal}
\end{figure}

A specific example for analysis is listed in Table \ref{A1:contaminated_text}. The corresponding SHAP values are shown in Figure~\ref{fig:shap_normal}~(a).
The tokens that have been identified as part of the contamination are among the ten most important features, together with rather general features like punctuation. 
In contrast, Figure~\ref{fig:shap_normal}~(b) shows the SHAP values for a model trained on the cleaned data and evaluated on the same text as in the previous experiment. Here, the set of tokens that influence the decision is more diverse. Additionally, the overall contribution of individual features to the prediction ``\gls{LLM}-generated'' is less pronounced.

A further experiment applies adversarial attacks to highlight the potential weaknesses of training detectors based on such data.
Once again, we compare models trained on the cleaned and uncleaned Claude multi-domain datasets.
We evaluate the classification performance using a separate test dataset. The evaluation data uses inputs labeled as ``human'', prefixed with the attack phrase \textit{'Here is a 7 sentence abstract for the provided article title: ’} in the beginning of the text.

Table \ref{tab:advers_attacks} shows that the attack is effective. It significantly decreasing the classification accuracy of the unadjusted model. Performance decreases from 99.9\% when classifying both human- and AI-written texts, to just 12.1\% when classifying the attacked human samples (Table \ref{tab:advers_attacks}). By contrast, cleaning the training data protects the model from such attacks, ensuring that the accuracy with which it classifies human-written samples remains at 97.1\% in attack scenarios.

\begin{figure}[tb]
    \centering
    \begin{subfigure}[t]{0.48\textwidth}
        \centering
        \includegraphics[width=\linewidth]{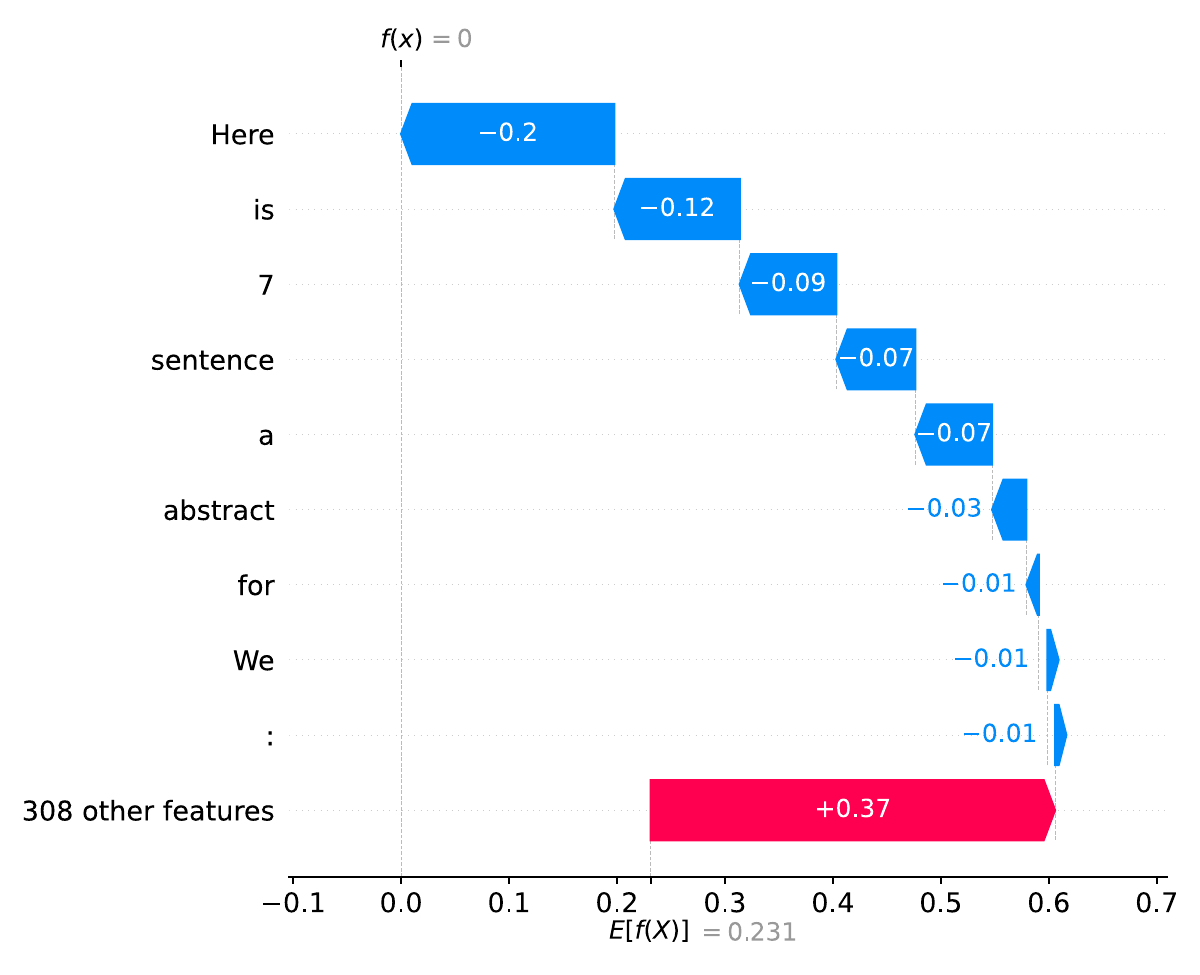}
        \caption{Model trained on uncleaned Claude data.}
    \end{subfigure}
    \hfill
    \begin{subfigure}[t]{0.48\textwidth}
        \centering
        \includegraphics[width=\linewidth]{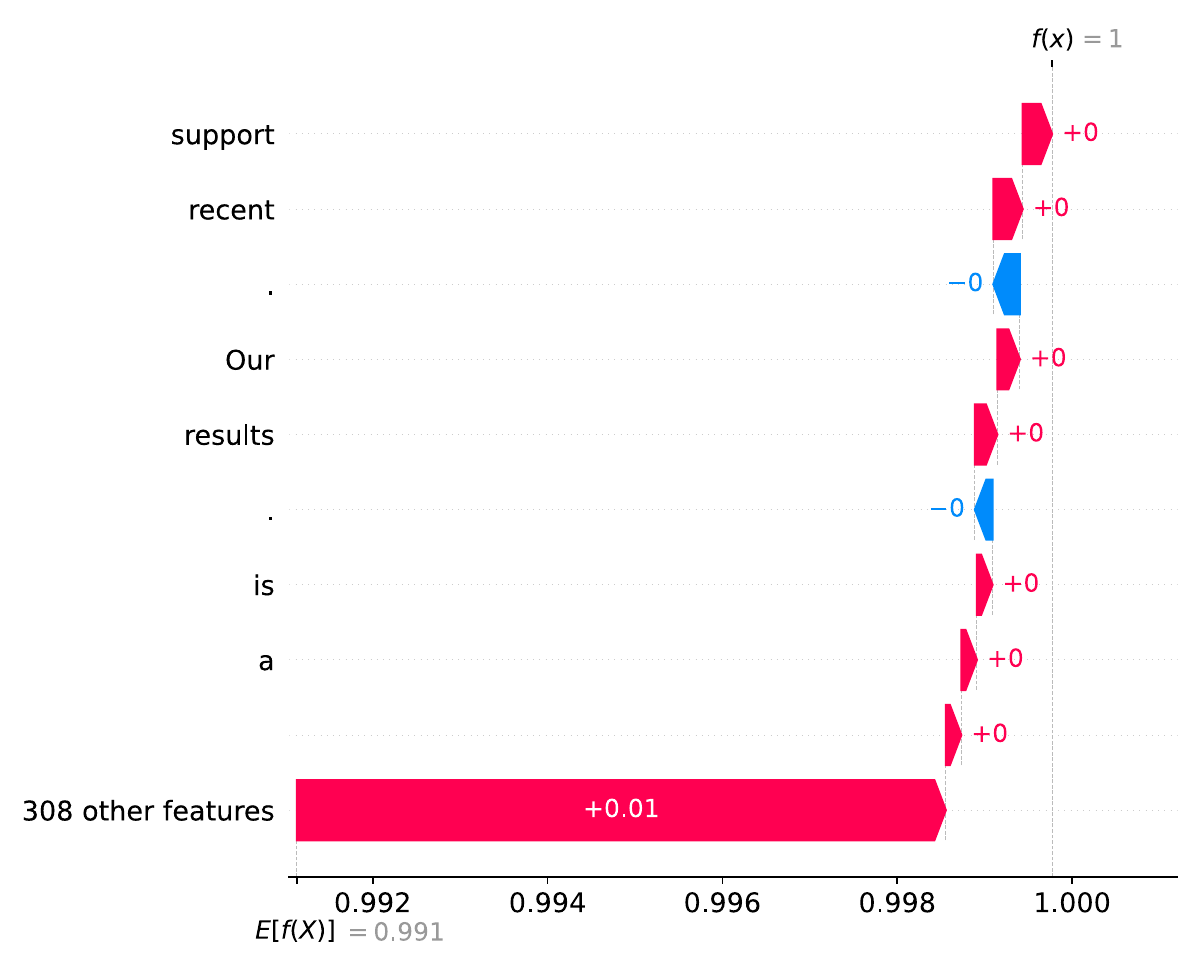}       
        \caption{Model trained on cleaned Claude data.}
    \end{subfigure}
    \caption{Comparison of SHAP token importance in a waterfall plot of a text for a model trained on
Claude-generated, multi-domain data. The sample is human-written and an adversarial attack is applied.}\label{fig:shap_attack}
\end{figure}

Figure \ref{fig:shap_attack} highlight the effectiveness of the attack on a concrete example.
The SHAP values show that the text would be classified as human-written if given only the other tokens. However, the attack heavily contributes to the class \gls{LLM}-generated, resulting in the sample being misclassified. However, a model trained on a cleaned dataset does not exhibit this behavior.

\begin{table}[tb]
\setlength{\tabcolsep}{4pt} 
\centering
\caption{Number of examples of human-generated data that is classified as 'Human' or 'AI-generated' when applying an adversarial attack.
The model was trained on Claude data in a multi-domain setting (test data size = 560).}
\label{tab:advers_attacks}
\begin{tabular}{p{0.5\textwidth} p{0.21\textwidth} p{0.21\textwidth}}
\hline
\textbf{} & 
\textbf{Model trained on original data} &
\textbf{Model trained on cleaned data} \\
\hline
\textbf{Prediction $=$ Human} & 68 & 544\\
\textbf{Prediction $=$ LLM} & 492 & 16 \\
\hline
\end{tabular}
\end{table}
\begin{table}[tb]
\centering
\caption{Generalisation performance across domains. Model trained on the cleaned dataset evaluated on cleaned and uncleaned data (and vice versa for an uncleaned training dataset). The contamination does not impact the performance on tasks from other domains.}
\label{tab:results_training}
\resizebox{\textwidth}{!}{
\begin{tabular}{p{3cm} p{1.5cm} cccc}
\hline
\textbf{domain\_train} & \textbf{cleaned} & \textbf{roc\_auc} & \textbf{f1} & \textbf{accuracy} & \textbf{tpr@0.01\%fpr} \\
\hline
arxiv & False & 0.878 & 0.803 & 0.807 & 0.182 \\
      & True  & 0.878 & 0.804 & 0.805 & 0.179 \\
\hline
arxiv\_cleaned & False & 0.846 & 0.782 & 0.785 & 0.163 \\
              & True  & 0.854 & 0.793 & 0.794 & 0.178 \\
\hline
\end{tabular}%
}
\end{table}

The findings raise the question of how representative the performance of the contaminated model would be compared to the performance of a model trained on cleaned data. 
This may impact scenarios such as plagiarism detection, where such tell-tale signs of AI-generated text could be easily be removed.
To analyse this, we trained two models on Arxiv data generated by Claude: one on the original samples and one on the cleaned samples.
We then evaluate the generalisation performance of the two \gls{roberta} models on the cleaned and uncleaned data from other domains and language models to determine whether focusing on contamination prevents the model from learning other features that are representative of the detection task.
Table \ref{tab:results_training} shows the results. The classifier's ability to  generalise to other domains is not impacted by the contamination. Apparently, both tasks — detecting the cleaned and uncleaned data — are comparably difficult for the classifier.

\section{Summary and Outlook}
Benchmarking of LLM text detectors is a difficult task. In this work, we examine the issue of contamination in the DetectRL benchmark. We show that the text generation artifacts impact the evaluation with DetectRL. 
At the same time, this contamination makes the detector susceptible to adversarial attacks. 
However, our initial results also show that the generalisation capabilities of a trained \gls{roberta} model are not affected by these artifacts in the data.
This article outlines our ongoing work to make users of the benchmark aware of the possible vulnerabilities.

We provide a reprocessed version of the dataset for download.
We cleaned the dataset primarily using regular expressions, mainly due to its massive size.
Despite the quality checks and multi-stage cleaning process, frequent artifacts or rejections by the LLM may still occur, as it was not possible for us to manually review all of the over 50,000 samples. However, we are confident that the proportion is considerably lower in the reprocessed dataset.

In future work, cleaning all samples using an LLM may have the potential to extend the cleaning process beyond the known artifacts and is less susceptible to minor changes in patterns.
More immediate next steps could be: first, to rerun the results of the DetectRL benchmark \cite{DBLP:conf/nips/WuZWY0YC24} on the cleaned data to identify potential differences.
Second, to regenerate the dataset using state-of-the-art LLMs, as the current ones seem outdated due to the rapid development of LLMs. We are confident that high data quality can be ensured from the outset by continuously checking and adjusting the prompts used.
Third, to extend the analysis to other benchmarks to determine whether the problem is specific to this one or more widespread.

\begin{credits}
\subsubsection{\ackname} This work was supported by Deutsche Forschungsgemeinschaft (DFG, German Research Foundation) as  part of the Research and Training Group 2475 ``Cybercrime and Forensic Computing'' 
(grant number 393541319/GRK2475/2-2024).

\subsubsection{\discintname}
The authors have no competing interests to declare that are
relevant to the content of this article.
\end{credits}
  \newline

\bibliographystyle{splncs04}

\end{document}